\documentclass[runningheads]{llncs}
\usepackage[T1]{fontenc}
\usepackage{amsmath}
\usepackage{amssymb}
\usepackage{booktabs}
\usepackage{marvosym} 

\usepackage{booktabs}    
\usepackage{xcolor}      
\usepackage{colortbl}    
\usepackage{subcaption}  
\usepackage{multirow}    
\usepackage{url}
\usepackage{placeins}

\usepackage{graphicx,verbatim}
\usepackage{array}

\begin{document}
\title{MIMM-X: Disentangeling Spurious Correlations for Medical Image Analysis}
\author{
Louisa Fay\inst{1,2}\textsuperscript{(\Letter)} 
\and Hajer Reguigui \inst{2} 
\and Bin Yang\inst{2} 
\and Sergios Gatidis\inst{3} 
\and Thomas K\"ustner \inst{1} 
}
\authorrunning{L. Fay et al.}

\institute{
Medical Image and Data Analysis, University Hospital of Tübingen, Germany \and
Institute for Signal Processing and System Theory, University of Stuttgart, Germany \and
Stanford University, Department of Radiology, Stanford, USA\\
\email{louisa.fay@med.uni-tuebingen.de}}

\maketitle              

\setcounter{footnote}{0}
\begin{abstract}

Deep learning models can excel on medical tasks, yet often experience spurious correlations, known as shortcut learning, leading to poor generalization in new environments. Particularly in medical imaging, where multiple spurious correlations can coexist, misclassifications can have severe consequences.
We propose MIMM-X, a framework that disentangles causal features from multiple spurious correlations by minimizing their mutual information. It enables predictions based on true underlying causal relationships rather than dataset-specific shortcuts.
We evaluate MIMM-X on three datasets (UK Biobank, NAKO, CheXpert) across two imaging modalities (MRI and X-ray). Results demonstrate that MIMM-X effectively mitigates shortcut learning of multiple spurious correlations. The code is publicly available\footnote{\url{https://github.com/lab-midas/MIMM-X}}.


\end{abstract}
\section{Introduction}
\label{sec:introduction}
Deep Learning (DL) has transformed medical imaging due its power of identifying patterns in the image that are not immediately visible to the human eye, resulting in remarkable success for segmentation, disease detection and diagnosis \cite{suganyadevi2022review}. However, medical datasets are inherently heterogeneous, influenced by factors such as scanners, acquisition protocols, and patient demographics. These factors introduce spurious correlations, leading DL models to rely on superficial patterns, known as shortcuts, instead of true causal relationships \cite{kumar2023debiasing,veitch2021counterfactual,sun2023right}. Consequently, models that perform well within a training domain often fail to generalize to new distributions \cite{cohen2020limits,pooch2020can}. For instance, a model trained on a cohort where a disease is more prevalent in male patients may associate the disease with male sex, yielding to biased predictions when applied to mixed-sex populations. In this scenario, instead of learning the true complex anatomical features indicative of the disease, the model exploits a demographic shortcut that leads to biased decision-making in patient care.
In response, causal representation learning aims to address this issue by disentangling causal relationships from spurious correlations to improve robustness under distribution shifts and counterfactual changes. 
Existing strategies include counterfactual augmentation \cite{xu2020adversarial,yao2022improving}, dataset rebalancing \cite{ando2017deep,sagawa2019distributionally}, adversarial training \cite{ganin2015unsupervised}, and dependence minimization \cite{muller2024benchmarking,fay2023avoiding}.

The previous Mutual Information Minimization Model (MIMM) \cite{fay2023avoiding} addressed shortcut learning by disentangling a primary task from a \textit{single} spurious factor by minimizing mutual information (MI). Yet, real-world medical datasets often contain multiple, entangled spurious correlations.
To this end, we propose MIMM-X, a framework that simultaneously disentangles and observes a primary task from multiple spuriously correlated factors with minimal computational overhead. We validate MIMM-X across three large-scale medical imaging datasets using two modalities (brain MRI from the German National Cohort (NAKO)\cite{nako} and UK Biobank (UKB)\cite{ukb}, and chest X-ray from CheXpert \cite{irvin2019chexpert}). We demonstrate that MIMM-X mitigates shortcut learning of the primary task and improves generalization across distribution shifts caused by (i) induced spurious correlations and (ii) factors that may naturally act as spuriously correlated factors, without modifying their distribution in our experimental setup.

\begin{figure}
    \centering
    \includegraphics[width=1\linewidth]{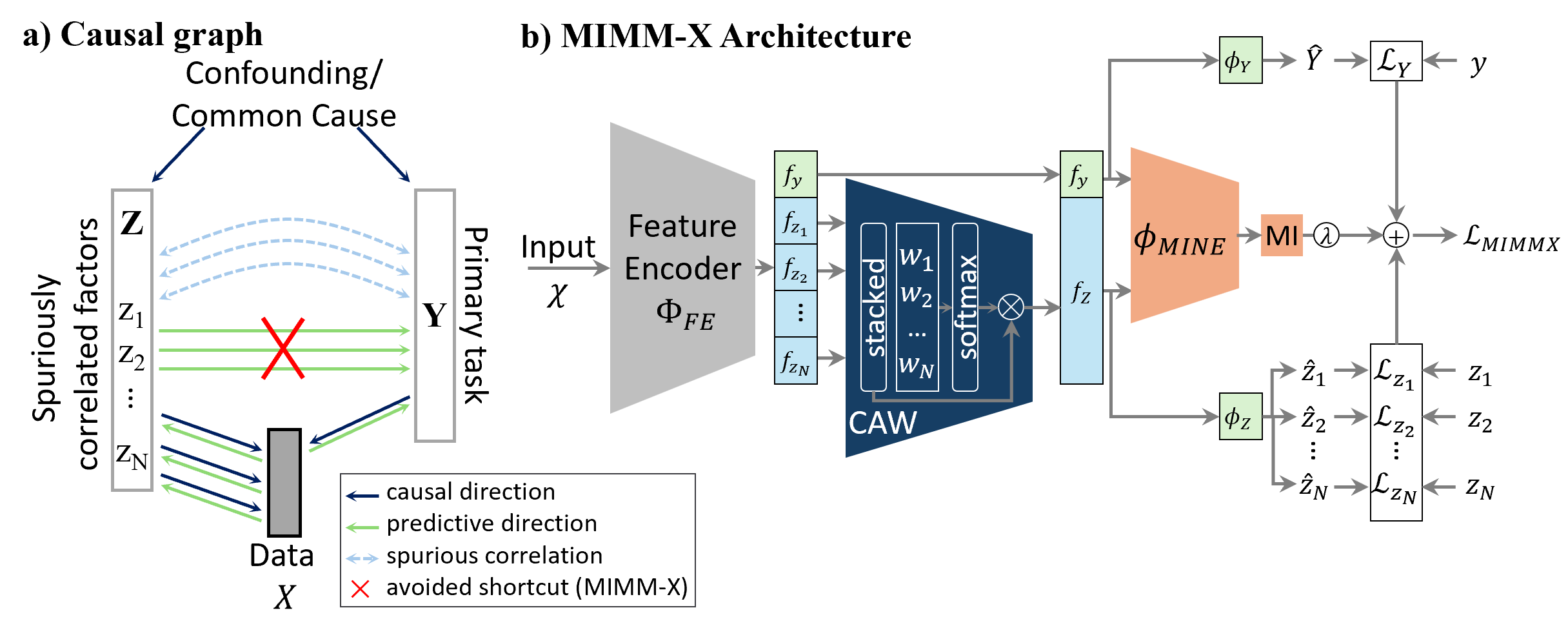}
    \caption{(a) Causal graph with multiple spurious correlations present. The aim is to avoid predictions based on shortcuts introduced by spurious correlations. (b) Our MIMM-X model promotes causal feature learning by minimizing the MI between the desired primary task $y$ and $N$ spurious correlations $Z$.
    }
    \label{fig:fig1}
\end{figure}

\section{Methods} \label{sec:methods}

We propose MIMM-X, designed to mitigate multiple spurious correlations $(z_i)_{i=1}^{N}$ while learning causal representations for a primary task $y$ from a vision input $X$ (Fig. \ref{fig:fig1}a). It is an extension of MIMM \cite{fay2023avoiding}, which, in comparison, can only handle a single known spurious correlation. As shown in Fig. \ref{fig:fig1}b, MIMM-X consists of a feature encoder $\phi_{FE}$, two classification heads $\phi_Y$ and $\phi_Z$, a mutual information estimator $\phi_{\text{MINE}}$, and a Confounder Attention Weighter module (CAW).

\subsection{Architectural Components}

\textit{Feature Encoder} $\phi_{FE}(X, \theta_{FE})$ maps an image $X$ to a feature vector $f$, partitioned into $N+1$ equal-sized subvectors. The subvector $f_y$ encodes the primary task, while $f_Z = [f_{z_1}, \dots, f_{z_N}]^T$ represents the $N$ spuriously correlated factors.

\textit{Classification Heads.} The primary task $y$ is predicted from $f_y$ through the classification head $\phi_Y$ with a log-softmax output. All $N$ spuriously correlated factors are predicted from the batch-normalized $f_Z$ with a linear multi-task classification head $\phi_Z$ and log-softmax.

\textit{Confounder Attention Weighter (CAW).} Before passing $f_z$ to the classification head $\phi_z$, CAW assigns adaptive attention weights to each spuriously correlated factor. A learnable parameter vector of size $N$ is normalized via softmax to produce attention weights $\mathbf{w} \in \mathbb{R}^N$. Each spuriously correlated feature subvector $f_{z_i}$ is then weighted by its corresponding attention weight $w_i$ to emphasize more relevant spurious factors. The attention weights are learned jointly with $\phi_{FE}$.

\textit{Mutual Information Estimation $\phi_{MINE}$.} To enforce independence between $f_y$ and $f_Z$, we minimize their mutual information (MI). We estimate a single MI value between $f_y$ and the stacked $f_Z$ using a MINE model \cite{belghazi2018mutual} to keep computational efficiency while scaling to multiple spurious correlations.

\subsection{Training Process}
MIMM-X uses alternating updates: $\phi_{FE}, \phi_{Y}, \phi_{Z}$ and \textit{CAW} are updated on one batch, followed by $N_B-1$ updates of $\phi_{MINE}$ where $N_B \in \mathbb{N}$ denotes the number of batches per iteration cycle.
The overall loss $\mathcal{L}_{\text{MIMM-X}}$ combines cross-entropy terms $\mathcal{L}_0$  for $y$ and $\mathcal{L}_{i}$ for  $i = 1,\ldots, N$ with MI penalty weighted by $\lambda$:
\begin{equation}
\mathcal{L}_{\text{MIMM-X}} = \gamma_{0}\mathcal{L}_{0} + \sum_{i=1}^{N} \gamma_{i} \mathcal{L}_{i} + \lambda  \phi_{MINE}(f_y, f_Z).
\end{equation}
Inspired by GradNorm \cite{chen2018gradnorm}, task-specific dynamic loss scaling (DLS) adjusts each task's contribution. Scaling factors $\gamma_i$ are defined as:
\begin{equation}
\gamma_i = \left(\frac{\mathcal{L}_i}{\overline{\mathcal{L}}}\right)^{\alpha_i}, \quad \text{where } \overline{\mathcal{L}} = \frac{1}{N+1} \sum\limits_{i=0}^{N} \mathcal{L}_i, \quad \text{and } \alpha_i = 
\begin{cases}
\alpha_Y, & \text{if } i = 0, \\
\alpha_Z, & \text{if } i = 1, \dots, N
\end{cases} \label{Eq2:DLS}
\end{equation}
with dynamic $\alpha_Y$ for the primary task:
\begin{equation}
\alpha_{Y} = \alpha_{Y, \text{initial}} + \left(\frac{\text{epoch}}{N_{\text{epoch}}} \cdot \beta_Y \right),
\end{equation}
where $\beta_Y$ controls the increasing emphasis on $y$ over training epochs.

\begin{figure}
    \centering
    \includegraphics[width=0.9\linewidth]{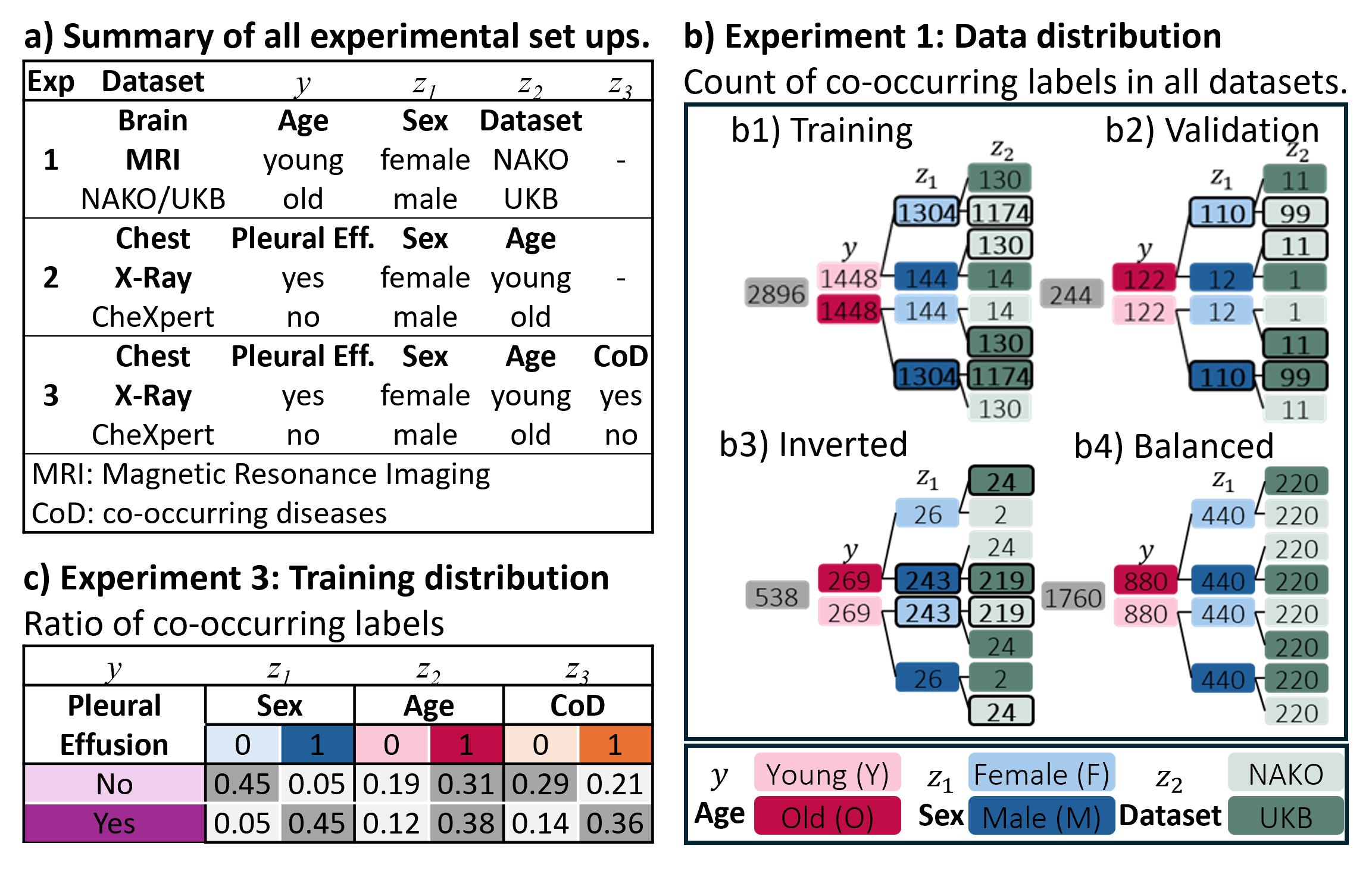}
   \caption{(a) Summary of experiments, primary tasks $y$ and spurious correlations $(z_i)_{i=1}^{N}$.
   (b) Experiment 1: Data distributions (absolute sample counts). (c) Experiment 3: Training data composition with synthetic and natural correlations.}
    \label{fig:dataset_distribution}
\end{figure}

\section{Materials}
We evaluated MIMM-X in three experiments with two modalities (Fig. \ref{fig:dataset_distribution}a).
\subsubsection{Training Setup.}
In Experiments 1–2, we introduced two strong correlations between $y$ and $(z_1, z_2)$ by sub-sampling training data. For each class of $y$, 90\% of samples were drawn from a specific class of $z_1$ and $z_2$, creating spurious correlations between $y$ and $z_1, z_2$ (see Fig.~\ref{fig:dataset_distribution}b1). Experiment 3 is based on one synthetic spurious correlation and additionally controls for two potentially naturally occurring spurious correlations (Fig.~\ref{fig:dataset_distribution}c). All experiments were run on a single NVIDIA GeForce RTX 3090 GPU.

\subsubsection{Evaluation.}
We evaluate all models on three types of datasets: \textit{Validation set (Val.)}: same spurious correlations as training (Fig. \ref{fig:dataset_distribution}b2); \textit{Inverted set (Inv.)}: inverse correlations (Fig. \ref{fig:dataset_distribution}b3); \textit{Balanced set (Bal.)}: no spurious correlation (Fig. \ref{fig:dataset_distribution}b4). If no performance drop is experienced when shifting from the validation distribution to the inverted test distribution, the model successfully avoided learning the shortcut introduced by the spurious correlation.

\textit{Evaluation against Comparison Methods.}
We compare MIMM-X to four methods. \textit{Baseline} uses the same architecture without the MI penalty. \textit{Rebalancing} balances the training set through resampling of under-represented spuriously correlated factor classes. \textit{Distance correlation (dCor)} \cite{szekely2007measuring,muller2024benchmarking} replaces MI with dCor as the dependence penalty. \textit{MIMM} \cite{fay2023avoiding} minimizes MI between the primary task and a single spurious factor; we trained separate models for $z_1$ and $z_2$, reporting the mean performance for $y$ and individual results for each $z_i$.

\textit{Evaluation of Disentanglements.}
We assess cross-predictive performance: using $f_y$ to predict $(z_i)_{i=1}^{N}$, and $f_{(z_i){_{i=1}^{N}}}$ to predict $y$ on the balanced test set. Ideal disentanglement yields random-guess performance, meaning that no information about the opposed task remains in the specific feature subvector.

\subsubsection{Experiment 1 (Brain MRI).}
We used central 2D axial slices from 3D brain MRI (NAKO/UKB), resized to $256 \times 256$ and z-score normalized. 
As primary task $y$, we predicted age as binary group (young: <51 years/old: >57 years) from the extracted slice, while we chose $z_1$: sex (female/male) and $z_2$: dataset (NAKO/UKB) as spuriously correlated factors.
We used 2,896 training, 244 validation, 538 inverted, and 1,760 balanced samples (Fig.~\ref{fig:dataset_distribution}a).
We applied the same feature encoder as in \cite{fay2023avoiding}, which is based on four convolutional layers. Training was performed using a batch size of 150 and $N_B=6$. DLS hyperparameters are set to $\alpha_{Y}=0.3$, $\alpha_{Z}=0.8$, $\beta_Y=0.01$, and $\lambda=1.5$.

\subsubsection{Experiment 2 (Chest X-ray).}
We used chest X-rays from CheXpert \cite{irvin2019chexpert} downsampled to $96\times96$ to predict pleural effusion (yes/no) as primary task $y$. The training set was strongly correlated by $z_1$: sex (female/male) and $z_2$: age (young: <50 years/old: >60 years).
Our training set included 4,126 X-rays, the validation 408 X-rays, the inverted test set of 2,692 X-rays, and the balanced set 4,432 X-rays. The feature encoder was a DenseNet-121 \cite{huang2017densely}, which was trained for 300 epochs with a learning rate of $10^{-5}$ and a batch size of 100 using $N_{B}=5$. The DLS hyperparameter were $\alpha_{Y}=0.3$, $\alpha_{Z}=0.7$, $\beta_Y=0.1$, and $\lambda=1.5$.

\subsubsection{Experiment 3 (Chest X-ray).}
We extended Experiment 2 to simulate real-world complexity.
We kept pleural effusion as primary task $y$ and introduced a single spurious correlation with sex ($z_1$) in the training set. Additionally, we controlled for two further factors that are often spuriously correlated with disease labels in clinical datasets: age ($z_2$) and presence of any of the remaining 12 co-occurring lung diseases (CoD, $z_3$) listed in \cite{chambon2024chexpert}.  These factors were not explicitly correlated with $y$. Their natural distributions were left unchanged to preserve clinically realistic associations. Importantly, $z_2$ and $z_3$ may still act as latent spurious correlates, potentially biasing predictions. This setup enables us to evaluate whether MIMM-X can disentangle $y$ not only from the known spurious factor $z_1$, but also from naturally co-occurring variables that are not manually manipulated. The training distribution is shown in Fig.~\ref{fig:dataset_distribution}c. The dataset included 9,850 training, 1,114 validation, 4,064 inverted, and 2,204 balanced test samples.


\begin{table*}[!htb]
\centering
\caption{\textbf{Experiment 1 (Brain MRI):} Primary task $y$: age (young/old); spuriously correlated factors $z_1$: sex (female/male), $z_2$: dataset (NAKO/UKB).}

\begin{subtable}{\textwidth}
\centering
\caption{\textbf{Classification accuracy [\%]} across different evaluation sets (Val./Inv./Bal.).}
\renewcommand{\arraystretch}{0.9}
    \begin{tabular}{@{}l|cc|ccc|ccc|ccc@{}}
        \toprule
        & & & \multicolumn{3}{c|}{\textbf{$f_y \rightarrow y$}} & \multicolumn{3}{c|}{\textbf{$f_z \rightarrow z_1$}} & \multicolumn{3}{c}{\textbf{$f_z \rightarrow z_2$}} \\

        \cmidrule(lr){4-6} \cmidrule(lr){7-9} \cmidrule(l){10-12}
        \textbf{Method} & \textbf{DLS} & \textbf{CAW} & \textbf{Val.} & \textbf{Inv.} & \textbf{Bal.} & \textbf{Val.} & \textbf{Inv.} & \textbf{Bal.} & \textbf{Val.} & \textbf{Inv.} & \textbf{Bal.} \\
        \midrule
        Baseline & -- & -- & 96.3 & 27.2 & 70.2 & 99.0 & 96.4 & 95.9 & 100.0 & 100.0 & 99.9 \\
        Baseline & \checkmark & \checkmark & 89.6 & 42.9 & 66.6 & 94.2 & 91.1 & 90.5 & 100.0 & 100.0 & 99.8 \\
        Rebalance & \checkmark & \checkmark & 91.8 & 81.9 & 87.3 & 97.5 & 96.7 & 97.3 & 100.0 & 100.0 & 100.0 \\
        dCor & -- & -- & 96.2 & 41.7 & 73.4 & 98.8 & 95.7 & 96.8 & 100.0 & 100.0 & 99.7 \\
        MIMM & -- & -- & 74.4 & 45.7 & 58.9 & 88.3 & 83.8 & 85.9 & 100.0 & 99.6 & 99.9 \\
        \midrule
        MIMM-X & -- & -- & 46.3 & 54.0 & 53.5 & 89.2 & 88.3 & 71.6 & 99.2 & 98.3 & 95.8 \\
        MIMM-X & -- & \checkmark & 94.2 & 33.2 & 69.0 & 63.3 & 57.5 & 66.5 & 54.6 & 44.5 & 57.3 \\
        MIMM-X & \checkmark & -- & 85.6 & 74.2 & 79.8 & 96.8 & 96.3 & 89.4 & 98.2 & 98.4 & 95.4 \\
        MIMM-X &\checkmark& \checkmark&84.2 & 82.8 & 82.6 & 93.4 & 94.6 & 93.6 & 100 & 99.8 & 99.5\\
        \bottomrule
    \end{tabular} \label{tab:exp1a}
\end{subtable}

\begin{subtable}{\textwidth}
\centering
\caption{\textbf{Disentanglement performance.} Accuracy in [\%]. Ideally, near random (50\%).}
\renewcommand{\arraystretch}{0.9}
\begin{tabular}{@{}l|ccc|ccc|ccc|ccc|ccc@{}}
\toprule
& \multicolumn{3}{c|}{\textbf{Baseline}} & \multicolumn{3}{c|}{\textbf{Rebalance}} & \multicolumn{3}{c|}{\textbf{dCor}} & \multicolumn{3}{c}{\textbf{MIMM}} & \multicolumn{3}{c}{\textbf{MIMM-X}} \\
& $y$ & $z_1$ & $z_2$ & $y$ & $z_1$ & $z_2$  & $y$ & $z_1$ & $z_2$& $y$ & $z_1$ & $z_2$  & $y$ & $z_1$ & $z_2$  \\
\cmidrule(lr){2-4} \cmidrule(lr){5-7} \cmidrule(lr){8-10} \cmidrule(lr){11-13}\cmidrule(l){14-16} 
$f_y$ & -- & 70.3 & 70.0 & -- & 47.6 & 48.1 & -- & 70.1 & 67.4 & -- & 40.4 & 41.4 & -- & 51.8 & 50.0 \\
$f_{z_1}$ & 51.7 & -- & 51.7 & 49.6 & -- & 49.4 & 52.3 & -- & 52.4 & 54.1 & -- & -- & 49.6 & -- & 51.5 \\
$f_{z_2}$ & 50.0 & 50.1 & -- & 50.1 & 50.0 & -- & 50.1 & 50.0 & -- & 50.7 & -- & -- & 49.8 & 50.1 & -- \\
\bottomrule
\end{tabular} \label{tab:exp1b}

\end{subtable}

\end{table*}

\section{Results and Discussion} \label{sec:results}

\textbf{Experiment 1.} \textit{Evaluation against Comparison Methods} (Table~\ref{tab:exp1a}). In this experiment, we aim to predict age group $y$ from brain MRI based on causal features, while avoiding shortcuts through induced spurious correlations: sex ($z_1$) and dataset origin, NAKO or UKB ($z_2$).
We first evaluated the impact of our DLS function (Eq.~\ref{Eq2:DLS}) and CAW module within MIMM-X. Although using CAW without DLS yielded the highest validation accuracy, it dropped by over -60\% on the inverted distribution, indicating reliance on spurious correlations ($z_1$, $z_2$) that no longer hold in this setting. We found that DLS dynamically balances the loss contributions, preventing the primary task signal from being dominated by easier-to-learn spurious correlations with $z_1$ and $z_2$. 

For the comparison methods, Baseline and dCor, performance drops by -61.6\% and -54.5\% from validation to inverted dataset. The previous model MIMM, designed for a single spurious correlation, also performs poorly in this multi-correlated setting, with performance of $y$ below random guessing on the inverted set. Rebalancing is more robust (-9.9\% drop).
However, MIMM-X with CAW and DLS effectively prevents shortcut learning and achieves the best generalization, with only a $1.4\%$ performance drop across distributions and the highest accuracy ($82.8\%$) on the inverted set. 
As expected, the prediction performance of the spuriously correlated factors remains high for all methods and datasets, as these tasks are inherently less complex to learn than age.

\begin{figure*}
    \centering
    \includegraphics[width=0.825\linewidth]{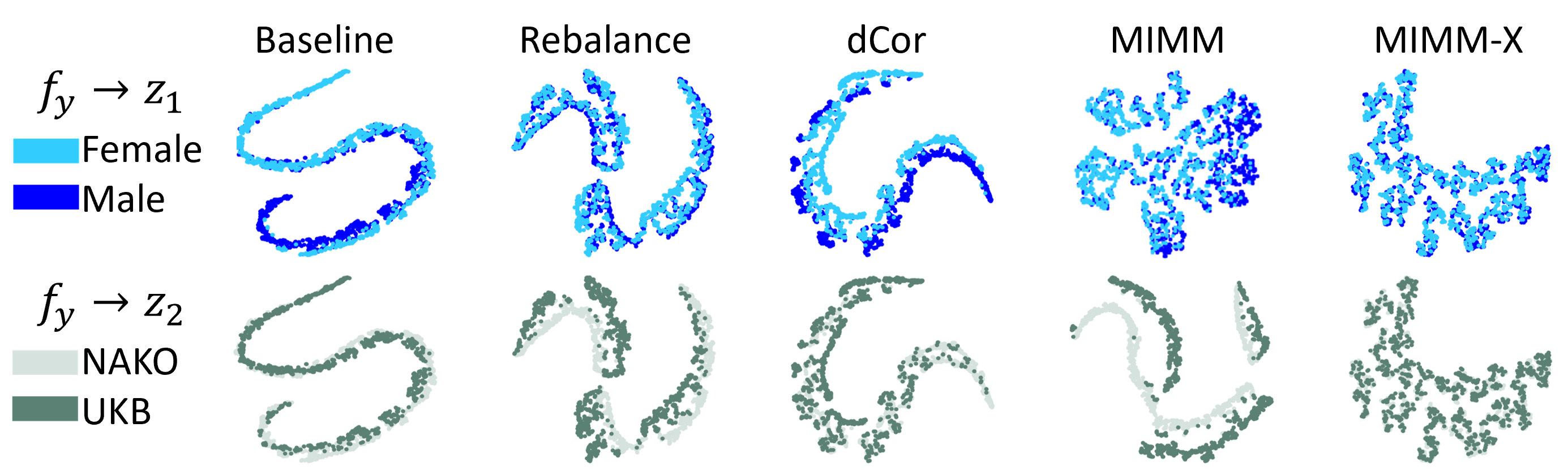}
    \caption{Experiment 1: t-SNE of $f_y$ colored by spurious factors $z_1$: sex, $z_2$: dataset. Ideally, $f_y$ is independent of $z_1$/$z_2$, meaning no visual class separation. While reference methods show clusters, MIMM-X is free from spurious information.} 
    \label{fig:tsne_nako_ukb}
\end{figure*}

\textit{Evaluation Disentanglement} (Table~\ref{tab:exp1b}). We evaluated disentanglement via cross-prediction: predicting $z_1$/$z_2$ from $f_y$ and $y$ from $f_{z_1}$/$f_{z_2}$ to assess residual shortcut information.
Only MIMM-X and Rebalancing achieved near random-guess performance when predicting $z_1$/$z_2$ from $f_y$, confirming effective disentanglement of causal and spurious features. However, compared to MIMM-X, Rebalancing increases the training sample size by 224.3\%, requiring more compute and training time. In contrast, Baseline and dCor retained substantial shortcut information. 
The t-SNE plots (Fig. \ref{fig:tsne_nako_ukb}) show that MIMM-X successfully disentangled $f_y$ from sex $(z_1)$ and dataset information $(z_2)$, as no clear separation is visible when coloring $f_y$ by the classes of $z_1$ (female/male) and $z_2$ (NAKO/UKB). Other reference methods still allow a separation based on spuriously correlated factors. 
These results highlight the effectiveness of MIMM-X in mitigating shortcut learning by better disentangling primary features of the age group from spurious correlations of sex and dataset.

\textbf{Experiment 2.} \textit{Evaluation against Comparison Methods} (Table~\ref{tab:exp2a}).

\begin{table*}[!htb]
\centering
\caption{\textbf{Experiment 2 (X-ray):} Primary task $y$: pleural effusion; spuriously correlated factors $z_1$: sex (female/male), $z_2$: age (young/old).}

\begin{subtable}{\textwidth}
\centering
\caption{Classification accuracy [\%] across different evaluation sets (Val./Inv./Bal.).}
\renewcommand{\arraystretch}{0.9}
\begin{tabular}{@{}l|ccc|ccc|ccc@{}} 
\toprule
& \multicolumn{3}{c|}{\textbf{$f_y \rightarrow y$}} & \multicolumn{3}{c|}{\textbf{$f_z \rightarrow z_1$}} & \multicolumn{3}{c|}{\textbf{$f_z \rightarrow z_2$}}\\
\cmidrule(lr){2-4} \cmidrule(lr){5-7} \cmidrule(l){8-10}
\textbf{Method} & \textbf{Val.} & \textbf{Inv.} & \textbf{Bal.} & \textbf{Val.} & \textbf{Inv.} & \textbf{Bal.} & \textbf{Val.} & \textbf{Inv.} & \textbf{Bal.} \\
\midrule
Baseline & 82.6 & 42.6 & 64.0 & 79.7 & 74.8 & 73.6 & 74.4 & 70.5 & 57.4 \\
Rebalance & 77.7 & 70.3 & 76.4 & 70.8 & 69.3 & 71.2 & 76.5 & 71.9 & 74.8 \\
dCor & 87.8 & 42.2 & 69.0 & 80.0 & 62.3 & 65.1 & 83.7 & 63.8 & 64.5 \\
MIMM & 87.2 & 51.0 & 71.4 & 72.1 & 68.9 & 66.7 & 63.6 & 64.1 & 53.1 \\
MIMM-X (ours) & 86.3 & 70.4 & 76.5 & 79.1 & 76.2 & 77.8 & 78.2 & 74.8 & 77.2 \\

\bottomrule
\end{tabular}\label{tab:exp2a}
\end{subtable}

\begin{subtable}{\textwidth}
\centering
\caption{Disentanglement performance. Accuracy in [\%]. Ideally, near random (50\%).}
\renewcommand{\arraystretch}{0.9}
\begin{tabular}{@{}l|ccc|ccc|ccc|ccc|ccc@{}}
\toprule
& \multicolumn{3}{c|}{\textbf{Baseline}} & \multicolumn{3}{c|}{\textbf{Rebalance}} & \multicolumn{3}{c|}{\textbf{dCor}} & \multicolumn{3}{c}{\textbf{MIMM}} & \multicolumn{3}{c}{\textbf{MIMM-X}} \\
& $y$ & $z_1$ & $z_2$ & $y$ & $z_1$ & $z_2$  & $y$ & $z_1$ & $z_2$& $y$ & $z_1$ & $z_2$  & $y$ & $z_1$ & $z_2$  \\
\cmidrule(lr){2-4} \cmidrule(lr){5-7} \cmidrule(lr){8-10} \cmidrule(lr){11-13}\cmidrule(l){14-16} 
$f_y$ & -- & 64.5 & 62.5 & -- & 54.9 & 53.9 & -- & 66.4 & 64.0 & -- & 61.1 & 61.0 & -- & 54.8 & 54.3 \\
$f_{z_1}$ & 57.9 & -- & 55.5 & 48.8 & -- & 49.4 & 56.8 & -- & 58.7 & 49.8 & -- & -- & 50.0 & -- & 50.0 \\
$f_{z_2}$ & 52.9 & 62.6 & -- & 52.7 & 50.5 & -- & 65.8 & 65.0 & -- & 50.4 & -- & -- & 49.6 & 50.1 & -- \\
\bottomrule
\end{tabular}\label{tab:exp2b}
\end{subtable}
\end{table*}

For our CheXpert experiment with two introduced spurious correlations ($z_1$:sex, $z_2$:age), we expected that shortcut learning is prevented if the performance on inverted and balanced test set for pleural effusion $y$ remained high. Similar to Experiment 1, we found that all comparison methods except for Rebalancing are near 50\% on the inverted distribution. Only Rebalancing and MIMM-X generalized well, with MIMM-X achieving the highest inverted accuracy (70.4\%).

\textit{Evaluation Disentanglement} (Table~\ref{tab:exp2b}). To evaluate disentanglement, we again predicted $y$, $z_1$, and $z_2$ from the opposed subvectors, i.e. $f_y \rightarrow z_1/z_2$; $f_{z_1} \rightarrow y/z_2 $; $f_{z_2} \rightarrow y/z_1 $.
Disentanglement results confirmed that MIMM-X and Rebalancing were closest to random guessing across all tasks, while other methods leave substantial spurious information in $f_y$ (accuracies >60\%).

\textbf{Experiment 3.}
\textit{Evaluation against Comparison Methods} (Table \ref{tab:exp3a}).
In this experiment, we investigated a more realistic setting, where we only induced one spurious correlation, $z_1$: sex, while $z_2$: age and $z_3$: CoD retained as naturally occurring correlations, i.e., their distribution in the training set was left unchanged (Fig. \ref{fig:dataset_distribution}c). We found that MIMM-X achieved highest performance on the inverted set with 70.2\% for the prediction of pleural effusion. Only Rebalancing achieved a similar performance, however, its performance on $z_1$ and $z_2$ is below MIMM-X. Other methods showed primary task performance below $<60\%$.

\textit{Evaluation Disentanglement} (Table \ref{tab:exp3b}). We evaluated disentanglement again by predicting the opposed task of the subvectors for the primary task and the three spuriously correlated factors. We found that $f_y$ of MIMM-X is closest to random guessing compared to the other comparison methods.

\begin{table}[!htb]
\centering
\caption{\textbf{Experiment 3 (X-ray):} Primary task $y$: pleural effusion; spuriously correlated factors $z_1$: sex (female/male), $z_2$: age (young/old), $z_3$: CoD (yes/no).}

\begin{subtable}{\textwidth}
\centering
\caption{Classification accuracy [\%] across different evaluation sets (Val./Inv./Bal.).}
\renewcommand{\arraystretch}{0.9}
\begin{tabular}{@{}l|ccc|ccc|ccc|ccc@{}}
\toprule
& \multicolumn{3}{c|}{\textbf{$f_y \rightarrow y$}} & \multicolumn{3}{c|}{\textbf{$f_z \rightarrow z_1$}} & \multicolumn{3}{c|}{\textbf{$f_z \rightarrow z_2$}} & \multicolumn{3}{c}{\textbf{$f_z \rightarrow z_3$}} \\
\cmidrule(lr){2-4} \cmidrule(lr){5-7} \cmidrule(lr){8-10} \cmidrule(l){11-13}
\textbf{Method} & \textbf{Val.} & \textbf{Inv.} & \textbf{Bal.} & \textbf{Val.} & \textbf{Inv.} & \textbf{Bal.} & \textbf{Val.} & \textbf{Inv.} & \textbf{Bal.} & \textbf{Val.} & \textbf{Inv.} & \textbf{Bal.} \\
\midrule
Baseline & 85.4 & 57.4 & 71.5 & 85.7 & 62.2 & 58.6 & 73.2 & 74.4 & 73.6 & 63.5 & 61.0 & 62.5 \\
Rebalance & 74.2 & 69.4 & 71.4 & 76.8 & 71.6 & 74.0 & 71.7 & 73.0 & 72.1 & 57.7 & 55.5 & 57.1 \\
dCor & 87.0 & 54.9 & 70.9 & 87.5 & 72.0 & 74.0 & 81.0 & 80.7 & 78.0 & 61.5 & 57.6 & 57.5 \\
MIMM-X (ours) & 81.7 & 70.2 & 72.9 & 81.5 & 72.7 & 74.2 & 81.4 & 78.3 & 79.4 & 58.9 & 57.5 & 58.1 \\
\bottomrule
\end{tabular}\label{tab:exp3a}
\end{subtable}

\begin{subtable}{\textwidth}
\centering
\caption{Disentanglement performance. Accuracy in [\%]. Ideally, near random (50\%).}
\renewcommand{\arraystretch}{0.9}
\begin{tabular}{@{}l|cccc|cccc|cccc|cccc@{}}
\toprule
& \multicolumn{4}{c|}{\textbf{Baseline}} & \multicolumn{4}{c|}{\textbf{Rebalance}} & \multicolumn{4}{c|}{\textbf{dCor}} & \multicolumn{4}{c}{\textbf{MIMM-X}} \\
& $y$ & $z_1$ & $z_2$ & $z_3$ & $y$ & $z_1$ & $z_2$ & $z_3$ & $y$ & $z_1$ & $z_2$ & $z_3$ & $y$ & $z_1$ & $z_2$ & $z_3$ \\
\cmidrule(lr){2-5} \cmidrule(lr){6-9} \cmidrule(lr){10-13} \cmidrule(l){14-17}
$f_y$ & -- & 66.6 & 60.9 & 60.3 & -- & 45.7 & 53.5 & 50.8 & -- & 70.0 & 62.2 & 61.3 & -- & 52.6 & 53.1 & 50.9 \\
$f_{z_1}$ & 65.9 & -- & 62.8 & 60.6 & 68.5 & -- & 62.8 & 58.5 & 62.6 & -- & 60.8 & 59.5 & 64.7 & -- & 59.9 & 56.0 \\
$f_{z_2}$ & 59.9 & 59.3 & -- & 57.1 & 57.1 & 55.0 & -- & 53.7 & 57.5 & 53.8 & -- & 52.5 & 56.9 & 53.5 & -- & 52.3 \\
$f_{z_3}$ & 55.4 & 53.9 & 60.1 & -- & 50.1 & 50.1 & 56.2 & -- & 50.0 & 50.0 & 56.0 & -- & 50.0 & 50.0 & 56.2 & -- \\
\bottomrule
\end{tabular}\label{tab:exp3b}
\end{subtable}

\end{table}

We acknowledge some limitations. MIMM-X currently relies on specifying confounding factors in advance. Extending it to discover unknown spurious correlations automatically and to handle more than three, as well as more complex confounding structures, will be explored in future work. Additionally, while dataset rebalancing can achieve comparable performance in some settings, it requires significantly larger training sets, increasing training time and computational cost. More advanced sampling strategies may mitigate this drawback, but they still rely on explicit knowledge of the spurious attributes and demand careful dataset design. In particular, the latter may not always be achievable in strongly non-homogeneous data cohorts. In contrast, MIMM-X offers a scalable solution that mitigates shortcut learning without modifying the data distribution.

\section{Conclusion}
This work presents MIMM-X, a scalable method for mitigating multiple spurious correlations and avoiding shortcut learning in DL-based medical image analysis. Our experiments demonstrate that MIMM-X improves robustness and feature disentanglement in scenarios with both induced and naturally occurring correlations, contributing toward causal and fair predictions.

\begin{credits}
\subsubsection{\ackname} 
    This project was conducted with data (Application No. NAKO-195 and NAKO-708) from the German National Cohort (NAKO) (www.nako.de). The NAKO is funded by the Federal Ministry of Education and Research (BMBF) [project funding reference no. 01ER1301A/ B/C and 01ER1511D], federal states of Germany and the Helmholtz Association, the participating universities and the institutes of the Leibniz Association. We thank all participants who took part in the NAKO study and the staff of this research initiative.

    This work was carried out under U.K. Biobank Application 40040. They also thank all participants who took part in the UKB study and the staff in this research program.

\subsubsection{\discintname}
The authors declare no competing interests.
\end{credits} 

%
%
%
\bibliographystyle{splncs04}
\bibliography{Paper-0019}

\end{document}